\documentclass[letterpaper, 10 pt, conference]{ieeeconf}  

\IEEEoverridecommandlockouts                              

\makeatletter
\let\NAT@parse\undefined
\makeatother

\usepackage[utf8]{inputenc} 
\usepackage[T1]{fontenc}    
\usepackage{hyperref}       
\usepackage{url}            
\usepackage{booktabs}       
\usepackage{amsfonts}       
\usepackage{nicefrac}       
\usepackage{microtype}      
\usepackage{color}
\usepackage[font=footnotesize,labelfont=bf]{caption}

\usepackage[dvipsnames]{xcolor}
\usepackage{ctable}
\usepackage{multirow}
\usepackage{wrapfig}
\usepackage{gensymb}
\let\labelindent\relax
\usepackage[inline]{enumitem}

\newcommand{\states}{\mathcal{S}}
\newcommand{\actions}{\mathcal{A}}

\newcommand{\stateinitial}{\rho_0}
\newcommand{\rewardfn}{r}
\newcommand{\transition}{\mathcal{T}}
\newcommand{\timemax}{T}
\newcommand{\policy}{\pi}

\newcommand{\mycheck}[1]{}

\newcommand{\cc}{\textcolor{black}}

\newcommand\blfootnote[1]{%
  \begingroup
  \renewcommand\thefootnote{}\footnote{#1}%
  \addtocounter{footnote}{-1}%
  \endgroup
}

\title{\LARGE \bf
Learning Dexterous Grasping\\with Object-Centric Visual Affordances
}

\author{Priyanka Mandikal$^{1}$ and Kristen Grauman$^{1,2}$
}

\begin{document}


\twocolumn[{%
 \renewcommand\twocolumn[1][]{#1}%
 \maketitle
\vspace*{-5.5mm}
 \begin{center}
  \centering
  \includegraphics[width=\linewidth]{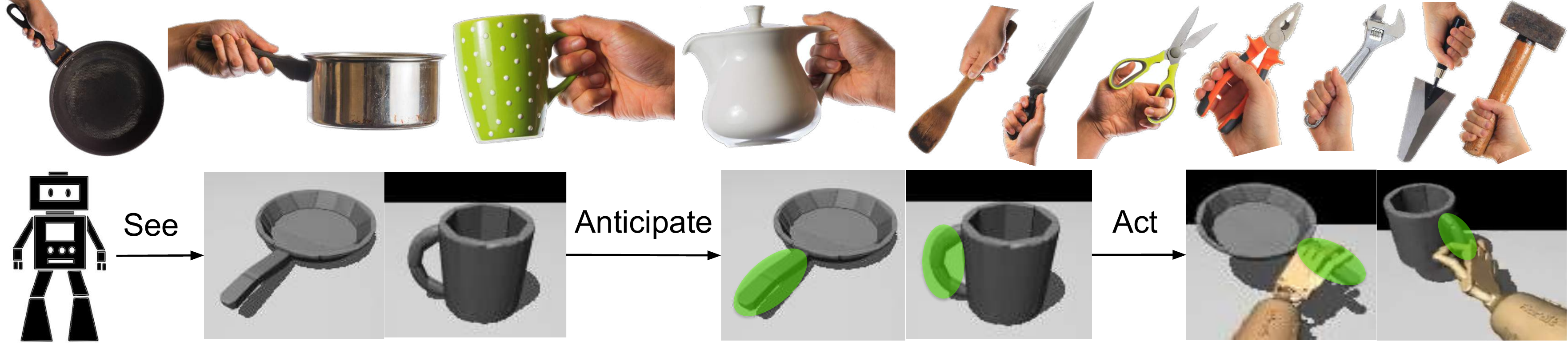}
 \captionof{figure}{\footnotesize{\textbf{Main idea.}  We aim to learn deep RL grasping policies for a dexterous robotic hand, injecting a visual affordance prior that encourages using parts of the object used by people for functional grasping. 
Given an object image (left), we
 predict the affordance regions (center), and use it to influence the learned policy (right).  The key upshots
 are better grasping, faster learning, and generalization to successfully grasp objects unseen during policy training.}}
  \label{fig:intro}
 \end{center}%
 }]

\begin{abstract}

Dexterous robotic hands are appealing for their agility and human-like morphology, yet their high degree of freedom makes learning to manipulate challenging. We introduce an approach for learning dexterous grasping. Our key idea is to embed an object-centric visual affordance model within a deep reinforcement learning loop to learn grasping policies that favor the same object regions favored by people.  Unlike traditional approaches that learn from human demonstration trajectories (e.g., hand joint sequences captured with a glove), the proposed prior is \emph{object-centric} and \emph{image-based}, allowing the agent to anticipate useful affordance regions for objects unseen during policy learning. We demonstrate our idea with a 30-DoF five-fingered robotic hand simulator on 40 objects from two datasets, where it successfully and efficiently learns policies for stable functional grasps. Our affordance-guided policies are significantly more effective, generalize better to novel objects, train 3$\times$ faster than the baselines, \cc{and are more robust to noisy sensor readings and actuation.} Our work offers a step towards manipulation agents that learn by watching how people use objects, without requiring state and action information about the human body. Project website with videos: \url{http://vision.cs.utexas.edu/projects/graff-dexterous-affordance-grasp}.

\end{abstract}
\blfootnote{$^1$ Department of Computer Science, The University of Texas at Austin
        }%
\blfootnote{$^2$ Facebook AI Research\newline
        {Correspondences to \tt\small mandikal@cs.utexas.edu}
        }%
\section{Introduction}

Robot grasping is a vital prerequisite for complex manipulation tasks. 
From wielding tools in a mechanics shop to handling appliances in the kitchen, grasping skills 
are essential to
everyday activity. 
Meanwhile, common objects are designed to be used by human hands (see Fig.~\ref{fig:intro}). Hence, there is increasing interest in dexterous, anthropomorphic robotic hands with multi-jointed fingers~\cite{gupta2016learning,rajeswaran2017learning,jain2019learning, zhu2019dexterous,akkaya2019solving,nagabandi2019deep,andrychowicz2020learning}.  Unlike simpler end effectors such as a parallel-jaw gripper, a dexterous hand has the potential for fine-grained manipulation.  
Furthermore, because its morphology agrees with that of the human hand, in principle it is readily compatible with the many real-world objects built for people's use.  Of particular interest is \emph{functional grasping}, where the robot should not merely lift an object, but do so in such a way that it is primed to use that object~\cite{brahmbhatt2019contactgrasp,kokic2020learning}.
For instance, picking up a pan by its base for cooking or gripping a hammer by its head for hammering is contrary to functional use.

Learning to perform functional grasping with a dexterous hand is highly challenging.  Typical hand models have 24 degrees of freedom (DoF) across the articulated joints, presenting  high-dimensional state and action spaces to master.  As a result, a reinforcement learning approach trained purely on robot experience faces daunting sample complexity.  Existing methods attempt to control the complexity by concentrating on a single task and object of interest (e.g., Rubik's cube~\cite{akkaya2019solving}) or by incorporating explicit human demonstrations~\cite{pathak2019,gupta2016learning,rajeswaran2017learning, jain2019learning,zhu2019dexterous,robobarista,state-only}.  For example, a human ``teacher" wearing a glove instrumented with location and touch sensors can supply trajectories for the agent to imitate~\cite{rajeswaran2017learning,jain2019learning,state-only}. While inspiring, this strategy is limited by its expense in terms of human time, the possible need to wear specialized equipment, and the close coupling between the person's arm/hand trajectory and the target object of interest, which limits generalization.

Towards overcoming these limitations, we propose a new approach to learning to grasp with a dexterous robotic hand.  Our key insight is to shift from \emph{person-centric} physical demonstrations to \emph{object-centric} visual affordances. Rather than learn to mimic the sequential states/actions of the human hand as it picks up an object, we learn the regions of objects most amenable to a human interaction, in the form of an image-based affordance prediction model. We embed this visual affordance model (a convolutional neural network) within a  deep reinforcement learning framework in which the agent is rewarded for touching the afforded regions. In this way, the agent has a ``human prior" for how to approach an object, but is free to discover its exact grasping strategy through closed loop experience.  Aside from accelerating learning, a critical advantage of the proposed object-centric design is generalization: 
the learned policy generalizes to unseen object instances because the image-based module can anticipate their affordance regions  (see Fig.~\ref{fig:intro}).

Our main contribution is to learn closed loop dexterous grasping policies with object-centric visual affordances.
We demonstrate our idea with the 30 DoF AdroitHand model~\cite{kumar2013fast} in the MuJoCo physics simulator~\cite{todorov2012mujoco}.  We train the visual affordance model from images annotated for human grasp regions~\cite{Brahmbhatt_2019_CVPR}. Importantly, image annotations are a much lighter form of supervision than state-action trajectories from today's status quo expert demonstrations.

In experiments with 40 objects, we show our approach yields significantly better quality grasps compared to other pure RL models unaware of the human affordance prior, even in the presence of sensor and actuation noise.  \cc{The learned grasping policies are stable under hostile external forces and robust to changes in the objects' physical properties (mass, scale).}  Furthermore, our approach significantly improves the sample efficiency of learning process, for a 3$\times$ speed up in training despite having no state-action demonstrations.  Finally, we show 
our agent generalizes to pick up object instances never encountered in training. 
For example, though trained to pick up a hammer, the model leverages partial visual regularities to pick up an axe. 
Our results offer a promising step towards agents that learn by \emph{watching} how people use real-world objects, without requiring information about the human operator's body.

\section{Related Work}

\noindent \textbf{Grasping with planning} 
Traditional analytical approaches use knowledge of the 3D object pose, shape, gripper configuration, friction coefficients, etc.~to determine an optimal grasp \cite{bicchi2000robotic,eigengrasps}.   
With the advent of deep neural nets, learning-based approaches to grasping have gained traction. 
A common protocol estimates the 6-DoF object pose, followed by model-based grasp planning~\cite{tremblay2018deep, mousavian20196, levine2018learning, ten2017grasp}.  Image modules trained to detect  successful grasps by parallel jaw grippers can accelerate the robot's learning~\cite{lenz2015deep,redmon,pinto2016supersizing,levine2018learning,mahler2017dex,murali2018cassl,qin2020s4g}.
The above strategies are typically employed for simple pick-up actions (not functional grasps) with simple end-effectors like parallel jaw grippers or suction cups, for which a control policy is easier to codify.  Some recent work explores related open-loop strategies with complex controllers, 
but, unlike our method, assumes access to the full 3D model of the objects~\cite{dogar2010grasping, amor2012generalization, bai2014dexterous, brahmbhatt2019contactgrasp, 8972562}.

\textbf{Reinforcement learning for closed-loop grasping}
Reinforcement learning (RL) models offer a counterpoint to the planning paradigm.  Rather than break the task into two steps---static grasp synthesis followed by motion planning---the idea is to use closed-loop feedback control based on visual and/or contact sensing so the agent can dynamically update its strategy while accumulating new observations~\cite{kalashnikov2018qt, quillen2018deep, merzic2019leveraging}.  Our proposed model is also closed-loop RL and hence enjoys this advantage.  However, unlike prior work, we inject an object-centric affordance prior learned from human grasps. It boosts sample efficiency, particularly important for the complex action space of dexterous robotic hands.  

Some impressive RL-based systems for dexterous manipulation tackle a specific task with a specific object, like solving Rubik's cube~\cite{akkaya2019solving}, shuffling Baoding balls~\cite{nagabandi2019deep}, or reorienting a cube~\cite{andrychowicz2020learning}.  In contrast, our focus is on grasping and lifting objects, including novel categories, and again our injection of object-centric human affordances is distinct.

\noindent \textbf{Learning manipulation with imitation} 
To improve sample complexity, imitation learning from expert demonstrations is frequently used, whether for non-dexterous~\cite{pathak2019,tcn,mime,robobarista} or dexterous~\cite{gupta2016learning, rajeswaran2017learning, jain2019learning, zhu2019dexterous,state-only} end effectors. 
Though advancing the state of the art in dexterous manipulation, the latter approaches rely on ``person-centric" human demonstrations with motion capture gloves.  
Aside from 
gloves, demonstrations may be captured via teleoperation and video~\cite{handa2019dexpilot} or paired video and kinesthetic demos~\cite{mime,pathak2019}.  In any case, expert demonstrations can be expensive, are specific to the end effector of the demonstration, and their trajectories need not generalize to novel objects.  

In contrast, the proposed object-centric affordances sidestep these issues, at the cost of instead supervising the predictive image model.  We use supervision from thermal image ``hotspots" where people hold objects to use them~\cite{Brahmbhatt_2019_CVPR}, though other annotation modes are possible.  ContactGrasp~\cite{brahmbhatt2019contactgrasp} leverages thermal image data to rank GraspIt~\cite{graspit} hand poses for a model-based optimization approach. 
In contrast, our approach 1) learns a closed-loop RL policy for grasping, and 2) incorporates a \emph{predictive} image-based affordance model that allows generalization to \emph{unseen} objects. 
Furthermore, once trained, our policy runs in real-time on new objects, whereas ContactGrasp takes about 4 hours to sample GraspIt poses for each unseen object.

\noindent \textbf{Visual affordances} 
A few methods infer visual affordances for grasping with simple grippers~\cite{redmon,kokic2017affordance,levine2018learning,kokic2020learning} and explore non-robotics affordances~\cite{myers2015affordance,affordancenet,nagarajan2019grounded,demo2vec}. Traditionally, supervision comes from labeled image examples~\cite{myers2015affordance,affordancenet} or a robot's grasp success/failure~\cite{redmon,lenz2015deep,levine2018learning}, while newer work explores weaker modes of supervision from video~\cite{nagarajan2019grounded,demo2vec}.
Recent work has shown that visual models can help focus attention for a pick and place robot~\cite{wu2020affordance,48887,zeng2018robotic}.
All of the prior methods make use of simple grippers in an open-loop control setting~\cite{redmon,lenz2015deep,levine2018learning,wu2020affordance,48887,zeng2018robotic,kokic2017affordance}.  To our knowledge, ours is the first work to demonstrate closed-loop RL policies learned with visual affordances.

\section{Approach}

Our goal is to learn dexterous robotic grasping policies influenced by object-centric grasp affordances from images. Our proposed model, called GRAFF for \emph{Grasp-Affordances}, consists of two stages (Fig.~\ref{fig:overview}). First, we train a network to  predict  affordance regions from static images (Sec.~\ref{sec:affordance_pred}). Second, we train a dynamic grasping policy using the learned affordances (Sec.~\ref{sec:policy_learning}). All of our experiments are conducted on a simulated tabletop environment using a 30 DoF dexterous hand as the robotic manipulator (detailed below). We next detail each of these stages.

\subsection{Affordance Anticipation From Images}
\label{sec:affordance_pred}

\begin{figure}[t]
\centering
\begin{center}
\includegraphics[width=0.85\linewidth]{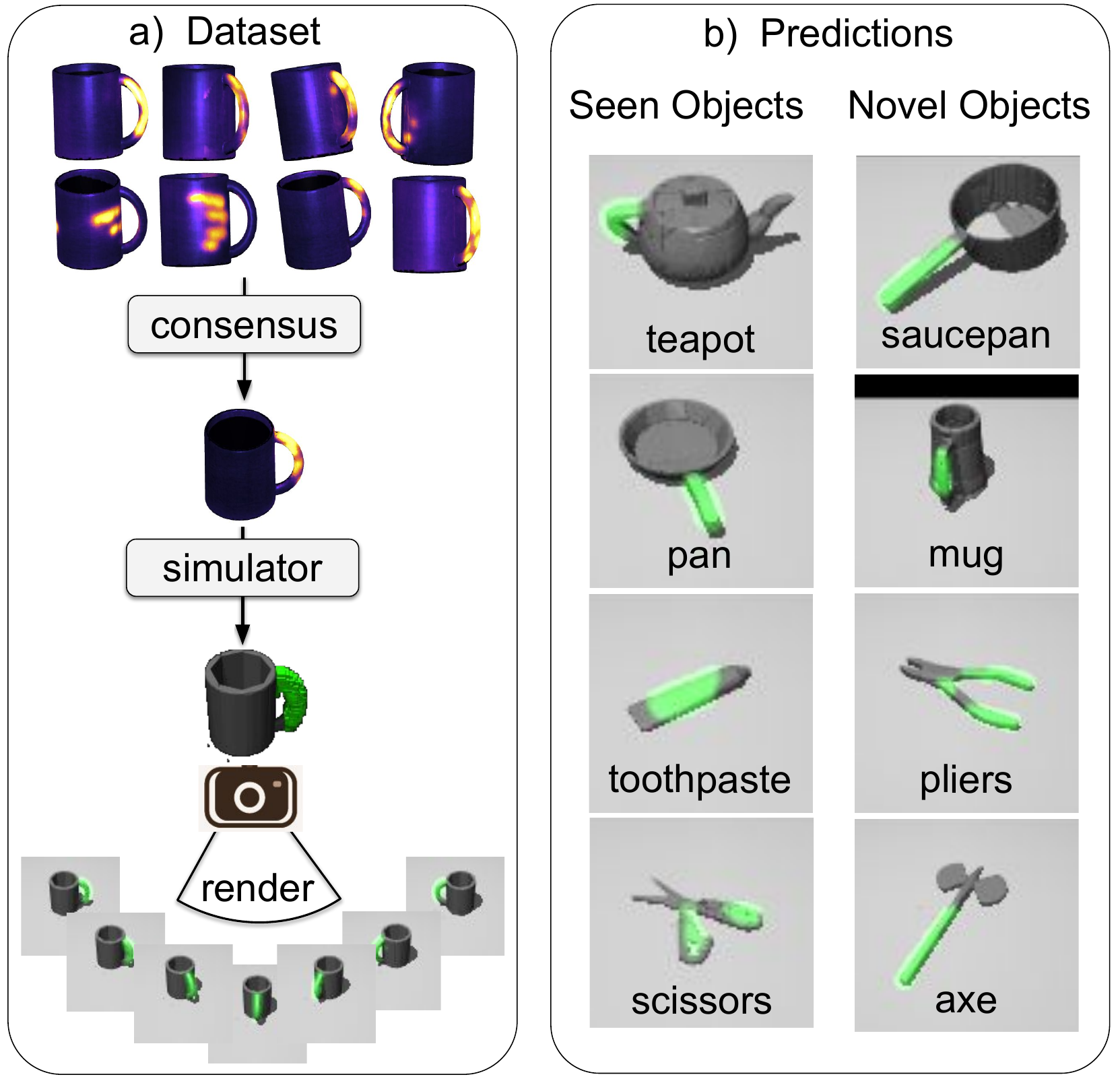}
\end{center}
\vspace*{-0.1in}
\caption{\textbf{Affordance anticipation.} a) Training images generated from 3D thermal maps from ContactDB.  Green denotes label masks overlaid on images. b) Sample predictions for seen and novel objects from ContactDB and 3DNet, respectively. Our anticipation model predicts functional affordances for novel objects and viewpoints (e.g., graspable handles and rings).}
\label{fig:affordance_figure}
\vspace{-0.5em}
\end{figure}

We first design a perception model to infer object-centric grasp affordance regions from static images. As discussed above, an object-centric approach has the key advantage of providing human intelligence about how to grasp while forgoing demonstration trajectories.  Furthermore, by predicting affordances from images, we open the door to generalizing to new objects the robot has not seen before.

\noindent \textbf{Thermal image contact training data}
We train the affordance model with images with ground truth functional grasp regions obtained from ContactDB~\cite{Brahmbhatt_2019_CVPR}.  
ContactDB contains 3D scans of 50 household objects along with real-world human contact maps captured using thermal cameras.
Participants grasped each object using two different post-grasp functional intents---\textit{use} and \textit{hand-off}---and a thermal camera on a turntable recorded the  multi-point ``hotspots" where the object was touched.   
Our model could alternatively be trained with manual image annotations. Note that our work \emph{infers} visual affordances on new images, whereas ContactDB is a dataset of actual grasp \emph{measurements}.

We consider contact maps corresponding to the \textit{use} intent and exclude objects having bimanual grasps, which yields 16 total objects.  Since each object has thermal maps captured from 50 different participants, we use  $k$-medoids clustering to obtain a representative thermal map for each object. Specifically, for a given object, we cluster the XYZ values of mesh points with a contact strength value above 0.5 (following~\cite{brahmbhatt2019contactgrasp}), then take the medoid of the largest cluster as our representative contact map for that object.
We port the 3D models into the MuJoCo physics simulator~\cite{todorov2012mujoco} and render them on a tabletop to create an image training set.

For each object, we obtain a set of image-affordance pairs $(x_i,y_i)$ by rendering the 3D object and the 3D contact map, respectively. See Fig.~\ref{fig:affordance_figure}a.
We rotate each object randomly within a 0-180$\degree$ range of the camera viewing angle and augment the dataset with varying camera positions.
Finally, we obtain a dataset of $\sim$15k training pairs, which we divide into an 80:10:10 train/val/test split.

\noindent \textbf{Image affordance prediction model} Let $X$ represent the domain of object RGB images, and let $Y$ be the object-centric grasp affordances. Our goal is to learn a mapping $G: X \rightarrow Y$ that will infer the grasp affordance regions from an individual image. 
During training, we have labelled (image, mask) pairs $\{x_i, y_i\}_{i=1}^N$. 
We pose the affordance learning problem as a segmentation task to predict binary per-pixel  labels, and approximate $G$ with a convolutional neural network. We adapt the Feature Pyramid Network (FPN)~\cite{lin2017feature} to perform semantic segmentation and use an ImageNet-pretrained ResNet-50~\cite{he2016deep} as the backbone. See Fig~\ref{fig:overview}a.

We now have a simple but effective model to infer object-centric grasp affordances from static images, which we will use below to guide a dexterous grasping policy.  
On the ContactDB test split, the segmentation accuracy averages 80.4\% in IoU.  Fig.~\ref{fig:affordance_figure}b shows sample predictions for both ContactDB and 3DNet~\cite{wohlkinger20123dnet} (
see Sec.~\ref{sec:exp} for dataset info).
Our affordance anticipation model is able to predict meaningful functional affordances for novel objects and viewpoints. For example, it faithfully infers graspable handles of saucepan, axe, and pliers despite not having encountered these categories in the training set.

\subsection{Dexterous Grasping using Visual Affordances}
\label{sec:policy_learning}

\begin{figure}[t]
\centering
\begin{center}
\includegraphics[width=\linewidth]{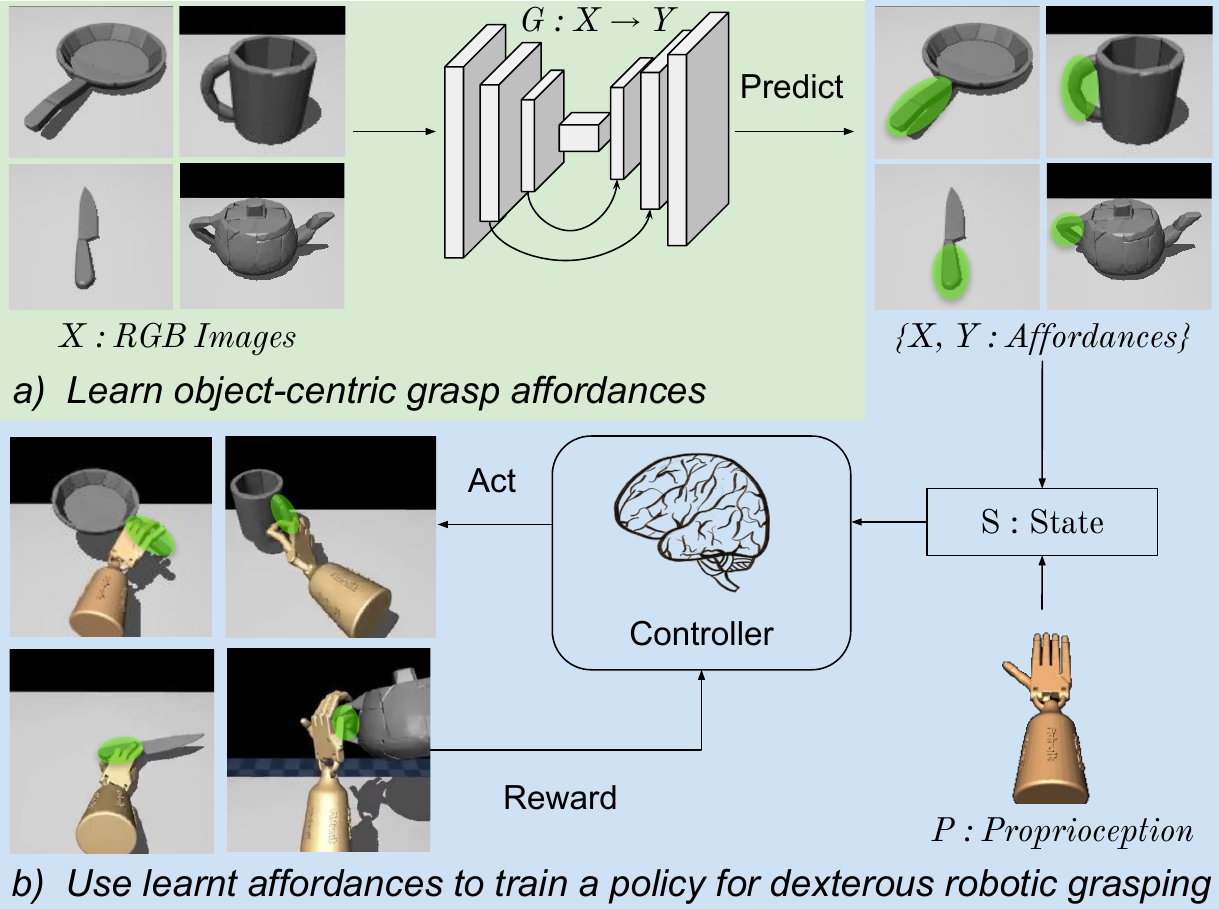}
\end{center}
\vspace*{-0.1in}
\caption{\textbf{Overview of our GRAFF model.} a) In Stage I, we train an affordance prediction model that predicts object-centric grasp affordances given an image. b) In Stage II, we train an RL policy that leverages these affordances along with other visuomotor sensory inputs (RGB-D image + hand joint variables) to learn a stable grasping policy.}
\label{fig:overview}
\vspace*{-1em}
\end{figure}

We want a controller that can intelligently process sensory inputs 
and execute successful grasps for a variety of objects with diverse geometries.
Towards this end, we develop a deep model-free reinforcement learning model for dexterous grasping.
Our robot model assumes access to visual sensing and proprioception, as well as 3D point tracking.  However, 
the agent does not have access to world dynamics, full object state, or the reward function.  
Given the large action and state spaces, sample efficiency is a significant challenge.  We show how the visual affordance model streamlines policy exploration to focus on object regions most amenable to grasping.
See Fig.~\ref{fig:overview}b.

\noindent \textbf{Problem formulation} We pose the problem of grasp acquisition as a finite-horizon discounted Markov decision process (MDP), with state space $\states$, action space $\actions$, state transition dynamics $\transition : \states \times \actions \to \states$, initial state distribution $\stateinitial$, reward function $\rewardfn : \states \times \actions \to \mathbb{R}$, horizon $\timemax$, and discount factor $\gamma \in (0, 1]$.
Hence, we are interested in maximizing the expected discounted reward 
$J(\policy) = \mathbb{E}_\policy [\sum^{\timemax-1}_{t=0} \gamma^t \rewardfn(s_t, a_t)]$
to determine the optimal stochastic policy $\policy : \states \to \mathbb{P}(\actions)$. We use an actor-critic model to estimate state values $V_{\theta}(s_t)$ and policy distribution $\policy_{\theta}(a_t|s_t)$ at each time step.

\begin{figure}[t]
\centering
\begin{center}
\includegraphics[width=\linewidth]{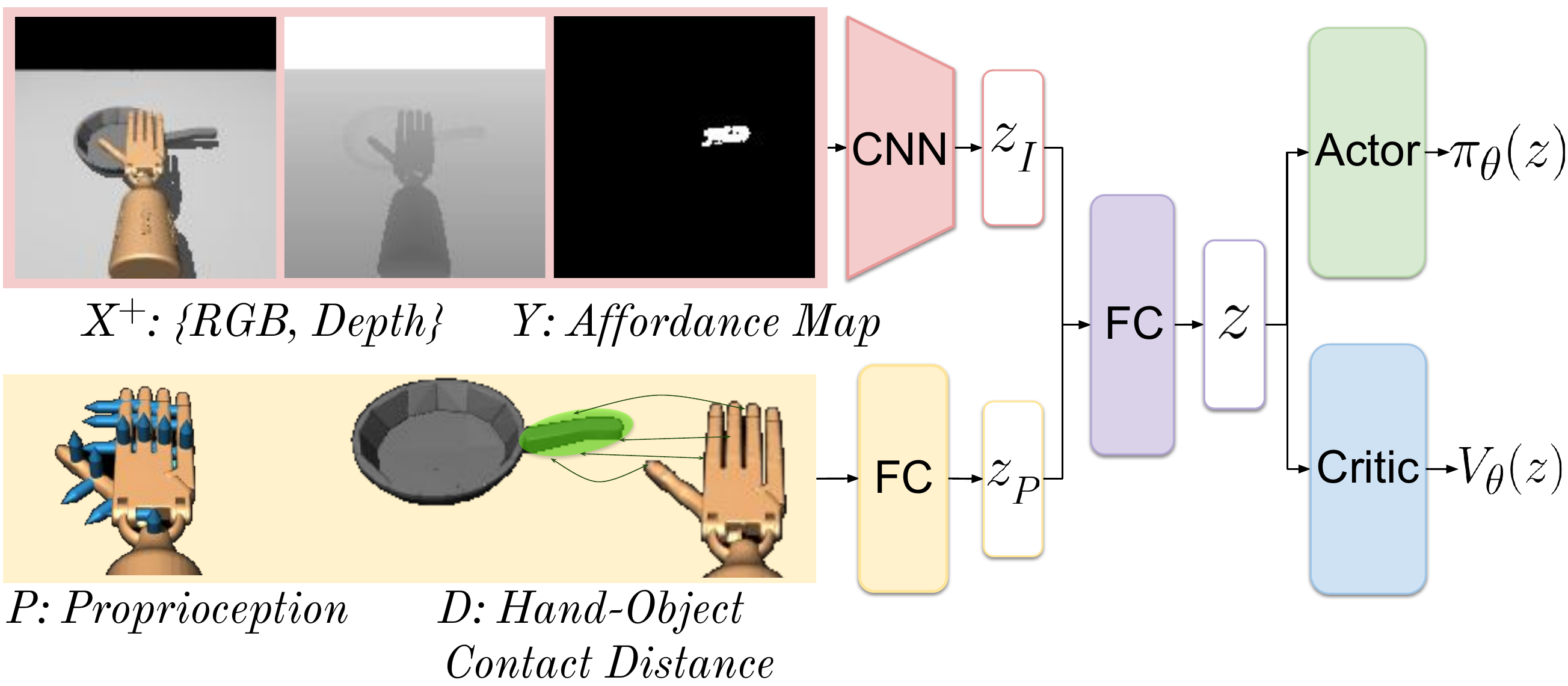}
\end{center}
\vspace*{-0.1in}
\caption{\textbf{Grasp policy learning architecture. } 
The inputs to the policy consist of RGB-D images and an inferred affordance map (top left), as well as the agent's proprioception and contact distances (bottom left). These streams are independently processed before being sent to an actor-critic network that samples actions and estimates state values.
} 
\label{fig:policy_learning}
\vspace{-1em}
\end{figure}

\noindent \textbf{State space} The work space of the robot consists of an object positioned on a table at a random orientation. 
The state space consists of the visuomotor inputs used to train the control policy: $\states = \{X^+,Y,P,D\}$ (see Fig.~\ref{fig:policy_learning}). 
The visual input at time $t$ consists of an RGB-D image $x_t^+ \in X^+$ captured by an egocentric hand-mounted camera that translates with the hand but does not rotate.
The affordance input $y_t \in Y$ is the binary affordance map inferred from the RGB image before the agent moves its hand in view, $y_t = G(x_t)$.  
The proprioception input $p_t \in P$ consists of the positions and velocities of each DoF in the hand actuator.

The distance input $d_t \in D$ is the distance between the agent's hand and the object affordance region.  We compute it as the pairwise distance between $M$ fixed points on the hand and $N$ points sampled from the backprojected affordance map.
We obtain the latter by backprojecting $y_0$ to 3D points in the camera coordinate system using the depth map at $t=0$, then tracking those points throughout the rest of the episode.  Hence we do not assume access to the full object state (we do not know the object mesh or mass), but we do assume perfect tracking of the affordance region that was automatically detected in the agent's first video frame. In experiments, we study the effect of substantial tracking failures to relax this assumption. 
We leave it as future work to incorporate SoTA visual tracking, e.g., by strengthening the segmentation model in the presence of occlusions.

\noindent \textbf{Action space} We use a 30-DoF position-controlled anthropomorphic hand from the Adroit platform \cite{kumar2013fast} as our manipulator. It consists of a 24-DoF five-fingered hand attached to a 6-DoF arm. Hence, our action space consists of 30 continuous position values. Delta angles are predicted by sampling from a multivariate Gaussian of unit variance whose mean is returned by the policy $\policy$.  

\noindent \textbf{Reward function}
The reward function should not only signal a successful grasp, but also guide the exploration process to focus on graspable object regions. 
To realize this, we combine two rewards: $R_{succ}$ (positive reward when the object is lifted off the table) and $R_{aff}$ (negative reward denoting the hand-affordance contact distance).
$R_{aff}$ is computed as the Chamfer distance between the $M$ and $N$ points described earlier.
We also include an entropy maximization term, $R_{entropy}$, to encourage exploration of the action space \cite{schulman2017proximal}.
Our total reward function is:
\setlength{\abovedisplayskip}{1em}
\setlength{\belowdisplayskip}{1em}
\begin{equation}
    \rewardfn = \alpha R_{succ} + \beta R_{aff} + \eta R_{entropy}.
\label{eq:reward}
\end{equation}
Through $R_{aff}$, the agent is incentivized to explore areas of the object that lie within the affordance region. The object-centric formulation poses no constraints on the hand pose, and can thus be seen as softer supervision than that employed in imitation learning for manipulation which requires kinesthetic teaching~\cite{mime, pathak2019} or tele-operation~\cite{handa2019dexpilot, rajeswaran2017learning, mandlekar2018roboturk}.

\noindent \textbf{Implementation details} We implement our approach with the architecture shown in Fig.~\ref{fig:policy_learning}.
The affordance network is optimized using Dice loss for 20 epochs with a learning rate of $1e-4$ and minibatch size of 8.
We preprocess the affordance map by computing its distance transform, which helps densify the affordance input.
The CNN encoder consists of three 2D convolutional layers with filters of size [8,4,3], and a bottleneck layer of dimension 512, with ReLU activations between each layer. The proprioception and hand-object distance inputs are processed using a 2-layer fully-connected encoder of dimension [512,512]. For the hand-object contacts, we use $M=10$ and $N=20$ uniformly sampled points. The CNN and FC embeddings are concatenated and further processed (FCs) before predicting the action values.  We optimize the network using the Adam optimizer~\cite{kingma2014adam} with a learning rate of $5e-5$. The full network is trained using PPO~\cite{schulman2017proximal}. 
We train a single policy for all ContactDB objects for 150M agent steps with an episode length of 200 time steps. With each episode being 2 s long, this amounts to $\sim$150 hours of learning experience. 
The coefficients in the reward function (Eq.~\ref{eq:reward}) are set as: $\alpha=1,\beta=1,\eta=0.001$. We train for four random seed initializations.
All project code is publicly available on the project website.
\section{Experiments}\label{sec:exp}

\noindent\textbf{Datasets} We validate our 
approach with two datasets: ContactDB~\cite{Brahmbhatt_2019_CVPR} and 3DNet~\cite{wohlkinger20123dnet}. 
We train a single policy across all 16 objects from ContactDB with one-hand grasps: apple, cell phone, cup, door knob, flashlight, hammer, knife, light bulb, mouse, mug, pan, scissors, stapler, teapot, toothbrush, toothpaste.  First we evaluate grasping on these 16 \textit{seen} objects. Then, we
test on 24 \emph{novel} object meshes from 3DNet,  a CAD model database with multiple meshes per category. We use 24 meshes from 9 categories that roughly align with the objects in ContactDB.
Four of the 3DNet categories exist in ContactDB (hammer, knife, mug, scissors), and the other five do not (axe, pencil, pliers, saucepan, wrench), making this a good test of generalization.

\noindent \textbf{Comparisons} We first devise two pure RL baselines 
that lack the proposed affordances:
\begin{enumerate*}[label=(\textbf{\arabic*})]
    \item \textsc{No Prior}: uses the lifting success and entropy rewards only. 
    \item \textsc{CoM}: uses the center of mass as a prior, which may lead to stable grasps~\cite{roa2015grasp,kanoulas2018center}, by penalizing the hand-CoM distance for $R_{aff}$. Both pure RL methods use our same architecture (Fig.~\ref{fig:policy_learning}), allowing apples-to-apples comparisons.
    \item \textsc{DAPG}: We also compare to DAPG~\cite{rajeswaran2017learning}, a hybrid imitation+RL model that uses motion-glove demonstrations.
    DAPG is trained with object-specific mocap demonstrations collected by us in VR for grasping each ContactDB object (25 demos per object). We stress that DAPG is a strongly-supervised approach with access to full motion trajectories of expert actions, whereas our approach uses inferred object-centric affordances to guide the policy. We train one policy per object for DAPG, allowing it to specialize to each object's demonstrations; our GRAFF is a single policy trained on all ContactDB objects. A practical advantage of our method is to replace heavy demonstrations (state-action pairs) with image-based affordances.
\end{enumerate*}

We stress that the task at hand is dexterous grasp \emph{acquisition} with a multi-fingered hand, not end effector pose estimation. 
Accordingly, we focus our comparisons on closed-loop RL methods to pinpoint exactly where our method has impact.  Non-RL methods that evaluate only end-effector pose, e.g.\cite{brahmbhatt2019contactgrasp}, as well as methods that directly regress 6-DoF grasp poses for a parallel-jaw gripper followed by grasp execution at that orientation~\cite{lenz2015deep,redmon,pinto2016supersizing,levine2018learning,mahler2017dex,murali2018cassl,qin2020s4g} are not applicable in this domain. In short, our idea is quite different---not only in approach (dynamic RL policy vs.~pose estimation), but also in problem domain (dexterous manipulation vs.~gripper) and learning signal (human use prior vs.~solely geometry or robot experience).

\noindent \textbf{Metrics}
We use three metrics:
\begin{enumerate*}[label=(\textbf{\arabic*)}]
    \item Grasp Success: For a given episode, a successful grasp has been executed if the object has been lifted off the table by the hand for at least the last 50 time steps (a quarter of the episode length) to allow time to reach the object and pick it up.
    \item Grasp Stability: After an episode completes, we apply perturbing forces of $1$ Newton in six orthogonal directions to the object. If the object remains held, the grasp is deemed stable.
    \item Functionality: We report the percentage of successful grasps in which the hand lies close to the GT affordance region (measured using Chamfer distance). 
\end{enumerate*}
We execute 100 episodes per object with the objects placed at different orientations ranging from [0,180\degree], and report mean and std dev of the metrics across all models trained with four random seeds.

\begin{figure}[t]
\centering
\begin{center}
\includegraphics[width=\linewidth]{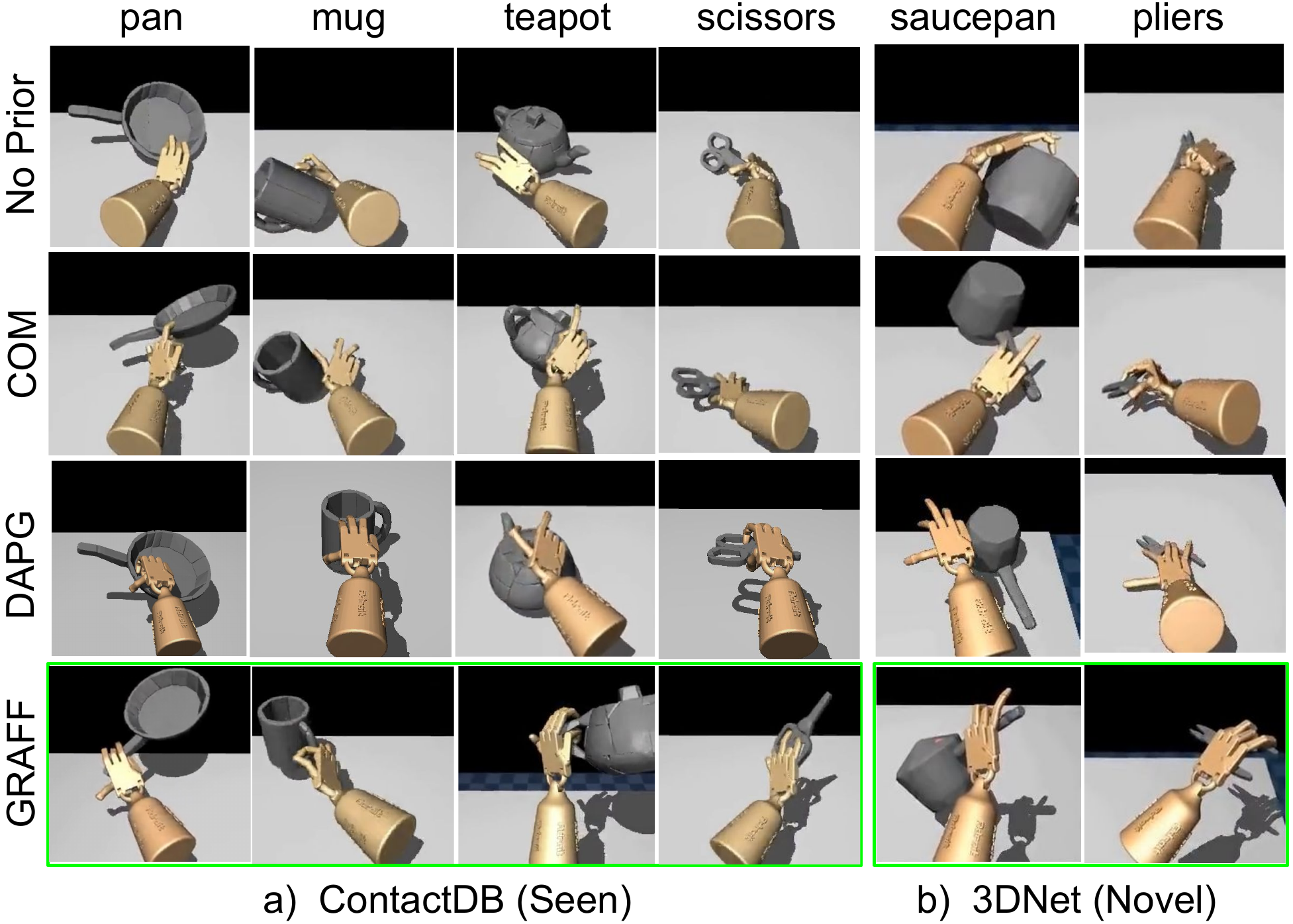}
\end{center}
\vspace*{-0.1in}
\caption{\textbf{Grasping performance.} Example frames from a) seen objects in ContactDB and b) novel objects in 3D-Net. Our affordance-based GRAFF is able to successfully grasp both seen and novel objects at their functional grasp locations, while the baselines either fail to learn successful grasps (mug, teapot, saucepan, pliers) or grasp at non-functional regions (pan, scissors). Despite GRAFF's weaker supervision compared to DAPG, it still generalizes better to unseen objects, thanks to the image-based model. 
}
\label{fig:grasp_results}
\vspace*{-1em}
\end{figure}
\begin{figure}[t]
     \centering
     \includegraphics[width=\linewidth]{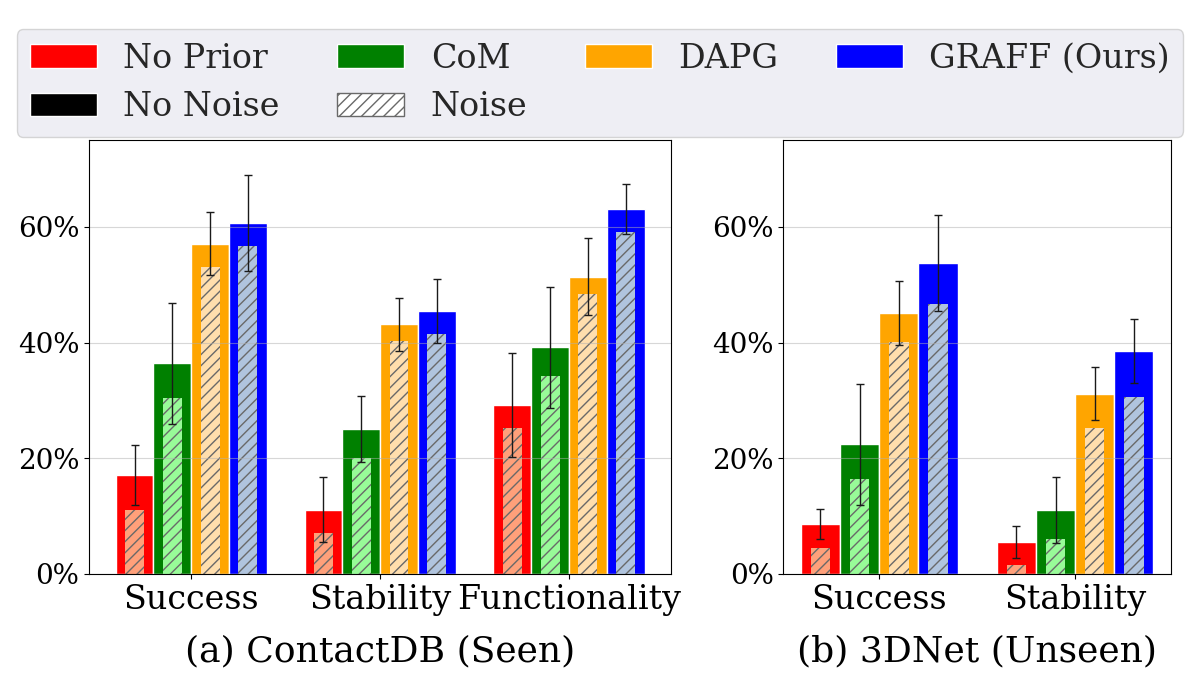}
     \vspace*{-0.15in}
     \caption{\textbf{Quantitative comparison}. Our approach outperforms the baselines and a state-of-the-art method that learns from expert demonstrations.  The differences remain even with noisy actuation, sensing, and tracking.}
    \label{fig:grasp_metrics}
\vspace{-0.5em}
\end{figure}

\noindent \textbf{Noisy sensing and actuation} When deploying trained policies on real robots, we might encounter a number of non-ideal circumstances owing to faulty sensor readings or imperfect robot executions. To better model such realistic settings, we incorporate noise into our training and testing regimes in simulation~\cite{akkaya2019solving,zhu2018reinforcement}. Following~\cite{zhu2018reinforcement}, we apply additive Gaussian noise on the proprioceptive sensor readings (robot joint angles and angular velocities), object tracking points, and robot actuation. We also apply pixel perturbations in the range [-5,5] and clip all pixel values between [0,255].
We train all versions of all methods under these noisy conditions (see Fig.~\ref{fig:grasp_metrics}, discussed below). We additionally test our method with a tracking failure model that freezes the track for 20 frames at random intervals and find that mean success rate is still 
reasonably high 
at 54\%. 
These empirical results indicate that GRAFF can remain fairly robust to 
tracking, sensing, and actuation failures that real robotic systems encounter.

While our lab does not have access to a real dexterous robotic hand, we believe that (like in~\cite{jain2019learning,li2020hierarchical,merzic2019leveraging,rajeswaran2017learning,xuegripping})
the high quality physics-based simulator together with these noise models offers a meaningful study.  
We are also encouraged by recent successes  
translating policies trained only in simulation to real-world dexterous robots using domain randomization~\cite{andrychowicz2020learning,tobin-iros2017}.

\noindent \textbf{Grasping seen objects from ContactDB} Fig~\ref{fig:grasp_metrics} shows the results on ContactDB. 
\cc{We show the mean and std dev over different seeds.}  
GRAFF (our model) outperforms both pure RL baselines consistently on all metrics.
Our gains persist with noisy sensing and actuation as well.
Fig.~\ref{fig:grasp_results}a shows qualitative examples; please see the video for full episodes and failure cases. GRAFF can successfully grasp the objects at the anticipated affordance regions (handle of pan, mug, teapot, knife, scissors), while the baselines fail to grasp objects with complex geometries (pan, mug, teapot). This shows the effectiveness of the affordance-guided policy in learning stable functional grasps. 

Our method also fares well compared to the more intensely supervised imitation+RL method DAPG~\cite{rajeswaran2017learning}, outperforming it on all metrics. This is a very encouraging result: not only does our method outperform its RL counterparts, but it is also competitive with a method that leverages expert trajectories. We found that DAPG can be vulnerable to imperfect demonstrations, yet in practice expert demos can be difficult to obtain (e.g., with mocap gloves). We also ran DAPG with the authors' provided demonstrations for a ball object, which performed slightly worse than the policies trained with the  object-specific DAPG policies in Fig.~\ref{fig:grasp_metrics}.

\begin{figure}[t]
     \centering
     \includegraphics[width=\linewidth]{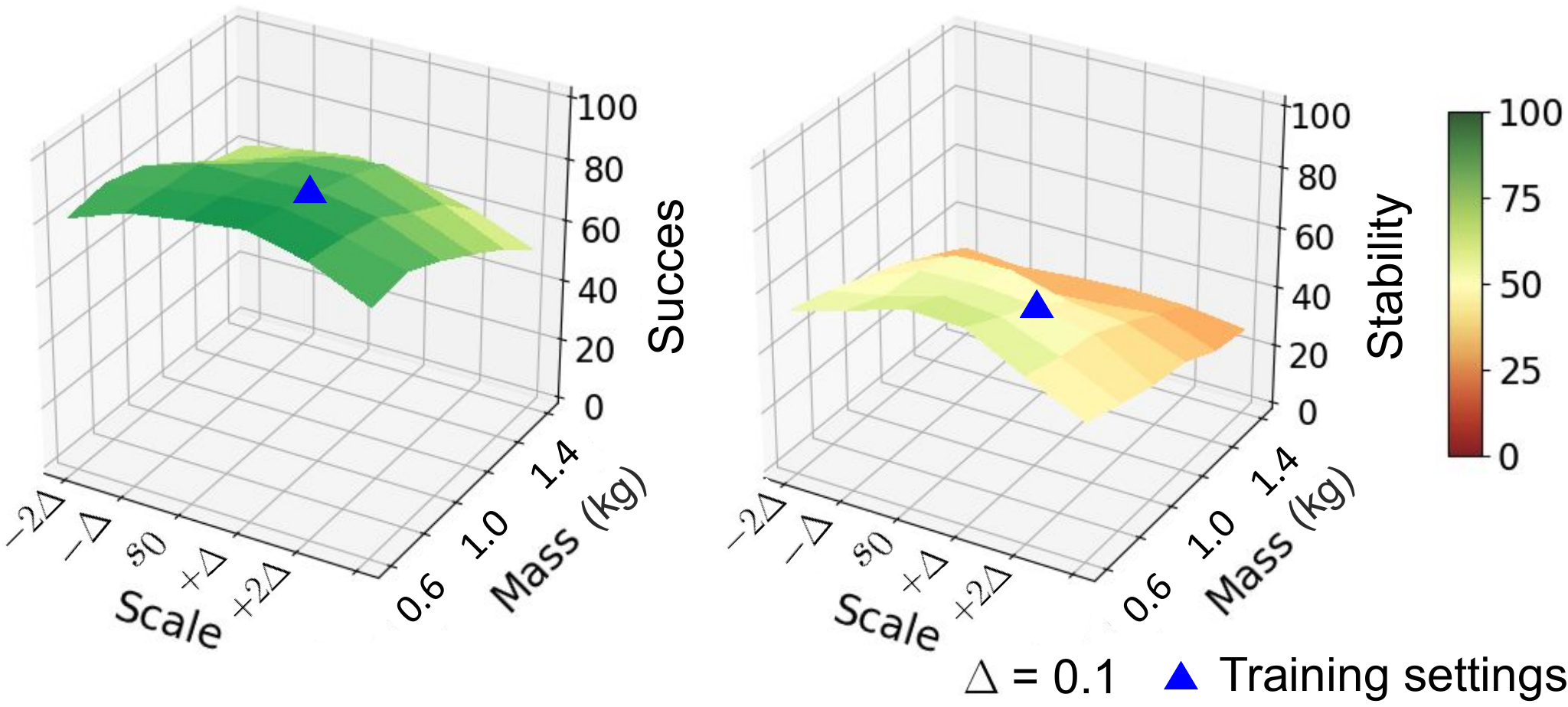}
     \vspace*{-0.15in}
     \caption{\textbf{Robustness to changes in physical properties}. 
    GRAFF shows good generalization across a range of mass and size variations of the objects.
     }
    \label{fig:generalization_mass_size}
\end{figure}

\noindent \textbf{Robustness to physical properties of the objects} To evaluate robustness to changes in object properties, we apply our policy to a range of object masses and scales not encountered during training. Fig.~\ref{fig:generalization_mass_size} shows 3D plots. 
Here, $m_0=1 kg$ and $s_0=1$ are the mass and scale values used during training. GRAFF remains fairly robust across large variations, which we attribute to GRAFF's preference for stable human-preferred regions. 

\noindent \textbf{Grasping unseen objects from 3DNet} Next we push the robustness challenge further by requiring the agent to generalize its grasp behavior to objects it has not encountered before
(the 24 3DNet~\cite{wohlkinger20123dnet} objects). We first render the objects and predict grasp affordances (cf.~Fig.~\ref{fig:affordance_figure}b). We then apply the policy trained on ContactDB to execute grasps. 
For DAPG, we use the trained policy of the ContactDB object that is closest in shape to the one in  3DNet.
Fig.~\ref{fig:grasp_metrics}b shows the results. We outperform all three baselines by a large margin in both grasp success and stability. 
The key factor is our visual affordance idea: the anticipation model generalizes sufficiently to new object shapes so as to provide a useful object-centric prior. 
Note that we cannot report the functionality score for 3DNet since there are no GT affordances for these objects.
Fig.~\ref{fig:grasp_results}b shows sample grasps. GRAFF successfully executes grasps at the anticipated affordance regions (e.g., handle of axe and finger rings on scissors), whereas the baselines may grasp the scissors at its blades or fail to lift the axe.

\begin{figure}[t]
     \centering
     \includegraphics[width=0.85\linewidth]{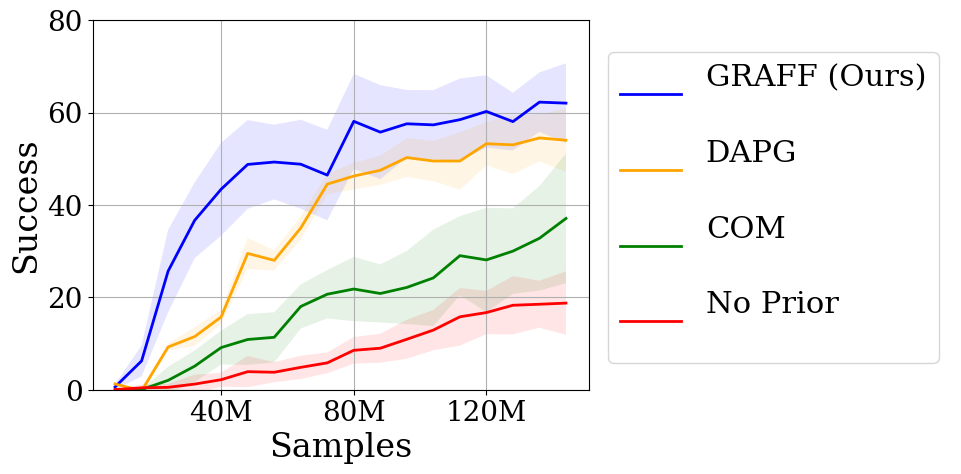}
     \vspace*{-0.05in}
     \caption{\textbf{Training curves.}  
     GRAFF outperforms both baselines by a large margin and provides a significant speed-up in terms of training time.
     }
     \label{fig:training_curves}
\vspace{-1em}
\end{figure}

\noindent \textbf{Ablations} The \textsc{No Prior} baseline above (average success rate 16\%) is a key ablation for our method.  To further tease out elements of our approach, we compare our model with only the hand-object contact term, adding the affordance map, and adding the $R_{aff}$ reward. Average success rates are 39\%, 42\%, 63\%, respectively. Thus our full model is most effective.

\noindent \textbf{Training time}  Fig.~\ref{fig:training_curves} shows the grasp success rate versus number of training samples.
Not only does our model learn more successful policies, it also has a sharper learning curve. While the pure RL baselines reach a maximum success rate of 30\% in 150M training samples ($\sim$150 hours of robot experience), our method reaches the same success rate in only 50M samples ($\sim$50 hours)---a 3$\times$ speedup.  Recall that 150 hours is a one-time cost: we train a single policy for all ContactDB objects, and simply execute that trained policy when encountering an unseen object. Thus our affordance prior meaningfully improves sample efficiency for dexterous grasping, while \cc{convincingly} outperforming the other pure RL methods. GRAFF also learns faster and performs better than the more heavily supervised DAPG for the same number of training samples.

\section{Conclusion}
 
Our approach learns dexterous grasping with object-centric visual affordances. Breaking away from the norm of expert demonstrations, our GRAFF approach uses an image-based affordance model to focus the agent's attention on ``good places to grasp". To our knowledge, ours is the first work to demonstrate closed-loop RL policies learned with visual affordances.  The key advantages of our design are its learning speed and ability to generalize policies to unseen (visually related) objects.
While there is much more work to do in this direction, we see the results as encouraging evidence for manipulation agents learning faster with more distant human supervision. In future work, we are interested in expanding the visual affordance model, modeling the multi-modal distribution of viable affordance regions, 
and investigating manipulations beyond grasping (e.g., open, sweep).

\textbf{Acknowledgements} UT Austin is supported in part by ONR PECASE N00014-15-1-2291.  Thanks to Samarth Brahmbhatt, Vikash Kumar, and Emo Todorov for helpful discussions.

\small
\bibliographystyle{plain}
\bibliography{egbib}

\end{document}